\documentclass[a4paper, 10pt, twocolumn, conference,comsoc]{IEEEtran}
\IEEEoverridecommandlockouts \IEEEpubid{\makebox[\columnwidth] 
	{\hfill 978-1-5386-6227-4/18/\$31.00~\copyright~2018 IEEE}
	\hspace{\columnsep}\makebox[\columnwidth]{ }}

\usepackage{amssymb}

\usepackage{graphicx} 

\usepackage{multirow}
\usepackage{booktabs}

\usepackage{xcolor}

\usepackage[colorlinks = true,
linkcolor = black,
urlcolor  = black,
citecolor = black,
anchorcolor = black,
bookmarks=false]{hyperref}

\ifCLASSINFOpdf
\else
\fi
\usepackage{array}
\usepackage{url}

\usepackage{etoolbox}

\makeatletter
\patchcmd{\@IEEEyesnumber}
  {\stepcounter}
  {\refstepcounter}
  {}{}
\patchcmd{\@@IEEEeqnarray}
  {\stepcounter}
  {\refstepcounter}
  {}{}
\patchcmd{\@@IEEEeqnarraycr}
  {\stepcounter{IEEEsubequation}}
  {\refstepcounter{IEEEsubequation}}
  {}{}
\patchcmd{\@@IEEEeqnarraycr}
  {\stepcounter{IEEEsubequation}}
  {\refstepcounter{IEEEsubequation}}
  {}{}
\patchcmd{\@@IEEEeqnarraycr}
  {\stepcounter{IEEEequation}}
  {\refstepcounter{IEEEequation}}
  {}{}
\patchcmd{\@@IEEEeqnarraycr}
  {\stepcounter{IEEEequation}}
  {\refstepcounter{IEEEequation}}
  {}{}
\makeatother

\usepackage[nameinlink]{cleveref}
\crefname{equation}{eq.}{eqs.}
\crefname{figure}{fig.}{figs.}

\begin{document}
%
\title{Benchmarking datasets for Anomaly-based Network Intrusion Detection: \mbox{KDD CUP 99} alternatives}



\author{\IEEEauthorblockN{Abhishek Divekar}
\IEEEauthorblockA{
\textit{Amazon}\\
Chennai, India \\
abhishek.r.divekar@gmail.com}
\and
\IEEEauthorblockN{Meet Parekh}
\IEEEauthorblockA{
\textit{New York University}\\
New York, USA \\
meetparekh09@gmail.com}
\and
\IEEEauthorblockN{Vaibhav Savla}
\IEEEauthorblockA{
\textit{Infosys}\\
Bangalore, India \\
vaibhav.savla95@gmail.com}
\and
\IEEEauthorblockN{Rudra Mishra}
\IEEEauthorblockA{
\textit{Samsung}\\
Bangalore, India \\
rudra.mishra35@gmail.com}
\and
\IEEEauthorblockN{Mahesh Shirole}
\IEEEauthorblockA{
\textit{Veermata Jijabai Technological Institute}\\
Mumbai, India \\
mrshirole@vjti.org.in}
}


%


\maketitle

\begin{abstract}
	
	Machine Learning has been steadily gaining traction for its use in Anomaly-based Network Intrusion Detection Systems (A-NIDS). Research into this domain is frequently performed using the KDD~CUP~99 dataset as a benchmark. Several studies question its usability while constructing a contemporary NIDS, due to the skewed response distribution, non-stationarity, and failure to incorporate modern attacks. In this paper, we compare the performance for KDD-99 alternatives when trained using classification models commonly found in literature: Neural Network, Support Vector Machine, Decision Tree, Random Forest, Naive Bayes and K-Means. Applying the SMOTE oversampling technique and random undersampling, we create a balanced version of NSL-KDD and prove that skewed target classes in KDD-99 and NSL-KDD hamper the efficacy of classifiers on minority classes (U2R and R2L), leading to possible security risks. We explore UNSW-NB15, a modern substitute to KDD-99 with greater uniformity of pattern distribution. We benchmark this dataset before and after SMOTE oversampling to observe the effect on minority performance. Our results indicate that classifiers trained on UNSW-NB15 match or better the Weighted F1-Score of those trained on NSL-KDD and KDD-99 in the binary case, thus advocating UNSW-NB15 as a modern substitute to these datasets.
	
\end{abstract}

\renewcommand\IEEEkeywordsname{Keywords}



%
\IEEEpeerreviewmaketitle

\section{Introduction}

Network security is an ever-evolving discipline where new types of attacks manifest and must be mitigated on a daily basis. This has led to the development of software to aid the identification of security breaches from traffic packets in real time: Network Intrusion Detection Systems (NIDS). These can be further categorized into misuse-based (M-NIDS) and anomaly-based systems (A-NIDS). M-NIDSs detect intrusions by an exact matching of network traffic to known attack signatures. An A-NIDS protects resources on computer networks by differentiating ordinary and suspicious traffic patterns, even those of which it has been previously unaware. Typically, A-NIDS systems are either statistical-based, knowledge-based or machine-learning based, with each category facing inherent drawbacks. Statistical techniques are susceptible to being gradually taught a deceptive version of normalcy, and rely on the quasi-stationary process assumption \cite{Patcha_et_al_overview_of_anomaly_detection_techniques_solutions_trends}. Knowledge-based expert systems employing predicates, state machine analysis \cite{Ilgun_et_al} and specifications \cite{Sekar_et_al_Specification_based_anomaly_detection_a_new_approach} suffer from scalability problems: as the number of attack vectors grows in tandem with the rise in volume and variety of network traffic, it becomes exceedingly difficult to construct an omniscient set of rules \cite{Teodoro_anomaly_based_nids_techniques_systems_challenges}. 

The state of art thus explores systems that automatically gain knowledge of how to distinguish benign usage patterns from malicious ones using a variety of machine learning techniques. In a machine learning approach, a classifier (or \textit{model}) is trained using a machine learning algorithm on a dataset of normal and abnormal traffic patterns. A trained model may subsequently be employed to flag suspicious traffic in real time. Such a dataset typically considers each pattern across several \textit{features} and an associated \textit{target class}, which denotes whether the pattern corresponds to normal or abnormal usage. Further training on fresh examples allows the model to adapt to the current network state. While systems adopting such an approach are not without fault, the preeminent advantages offered include the ability to gain knowledge without explicit programming, and adaptability to dynamic traffic patterns \cite{Patcha_et_al_overview_of_anomaly_detection_techniques_solutions_trends}.

The selection of a training dataset is integral to the security of a modern A-NIDS using machine learning techniques. In the ideal case, such datasets would be specific to each network deployment \cite{Outside_The_Closed_World}; however, a lack of alternatives has led to several works focusing on the \textit{KDD~CUP~99} dataset \cite{KDD_99_dataset} as a popular benchmark for classifier accuracy \cite{Shamshirband_et_al_wireless}. Unfortunately, KDD-99 suffers several weaknesses which discourage its use in the modern context, including: its age, highly skewed targets, non-stationarity between training and test datasets, pattern redundancy, and irrelevant features. Other researchers have devised several countermeasures that mitigate these flaws. An important effort by Tavallaee et al. \cite{Tavallaee_et_al} was to introduce \textit{NSL-KDD}, a more balanced resampling of KDD-99 where focus is given to examples which are likely to be missed by classifiers trained on the basic KDD-99. 

A significant trend that has been discovered is the poor performance of classifiers on minority classes of KDD-99, an obstacle which NSL-KDD is unable to eliminate. We argue that  this is due to the extreme skewedness of the dataset with respect to its target class distribution. In this work, we empirically prove that \textit{NSL-SMOTE} - a balanced dataset constructed using oversampling and undersampling - shows significant performance improvements over NSL-KDD on the same minority classes.

The inherent age of the dataset is another major drawback. At the time of this writing, KDD-99 is nearly two decades old and proves insufficient and archaic in the modern security context \cite{Outside_The_Closed_World}. We believe that the inertia of extant research using KDD-99 as the primary dataset makes authors reluctant to ``start from scratch" on another one when constructing complex algorithms. Hence, a major contribution of this work is to benchmark the performance of a modern and balanced dataset, \textit{UNSW-NB15}, on standard machine learning classifiers used widely in academia. Consequently, we observe the class-wise and Weighted F1-Score of Neural Network, Support Vector Machine, Decision Tree, Random Forest and K-Means classifiers trained on UNSW-NB15. We then contrast its performance to that of \textit{NB15-SMOTE}, an oversampling of UNSW-NB15's minority classes. Finally, binarization of targets eliminates imbalance and allows a direct comparison of the datasets. Extant research conducted on KDD-99 is thus aided by proving the suitability of basic machine learning techniques on a more rounded dataset.

The structure of this paper is as follows:  \Cref{data_desc} briefly describes the characteristics of KDD, NSL-KDD, and UNSW-NB15. In  \Cref{methodology} we present the Machine Learning pipeline that we apply to train classification models on each dataset. This  also discusses SMOTE oversampling, which is used to create the \textit{NSL-SMOTE} and \textit{NB15-SMOTE} datasets.  \Cref{results} showcases and analyzes the results of different classifiers obtained on these datasets. We conclude our paper with our inferences in  \Cref{discussion}.

\section{Dataset Description} \label{data_desc}

\subsection{KDD CUP 99 Dataset} \label{KDD_desc}
This is a modification of the dataset that originated from an IDS program conducted at MIT's Lincoln Laboratory, which was evaluated first in 1998 and again in 1999. The program, funded by DARPA, yielded what is often referred to as the \textit{DARPA98} dataset. Subsequently, this dataset was filtered for use in the International Knowledge Discovery and Data Mining Tools Competition \cite{Stolfo_et_al_KDD_creation}, resulting in what we recognize as the \textit{KDD CUP 99} dataset \cite{KDD_99_dataset}. 

McHugh \cite{McHugh_DARPA} noted several issues with DARPA98, including the unrealistic network architecture, overt synthesis of data, high tolerance for false alarms, and questionable evaluation methodology. Tavallaee et al. \cite{Tavallaee_et_al}, while proposing an improved \textit{NSL-KDD}, provided a comprehensive description of the KDD~CUP~99's idiosyncrasies. The pertinent aspects are noted below.

\subsubsection{Genesis}
This dataset constitutes the \textit{TCPdump} data of simulated network traffic captured in 1998 at Lincoln Labs. Seven weeks of traffic resulted in five million connection records used for a training set. A further two weeks of network traffic generated a test set with two million examples. The complete schedule can be found at the MIT website \cite{KDD_DARPA_attack_description_and_schedule}. KDD-99 is a filtered version of this data.

\subsubsection{Target classes}
KDD-99 has five classes of patterns: Normal, DoS (Denial of Service), U2R (User to Root), R2L (Remote to Local) and Probe (Probing Attack). Each intrusion category is further subclassified by the specific procedure used to execute that attack. We list the distribution of patterns into target classes in  \Cref{kdd_data_distribution}.

\subsubsection{Size and redundancy} \label{kdd_size_and_redundancy}
As provided, KDD's training dataset contains 4,898,431 data points. However, due to high redundancy (78\%), there exist only 1,074,992 unique data points. Similarly, the test dataset is highly redundant (89.5\%); it was pruned from 2,984,154 to 311,029 patterns. We consider these reduced datasets in this paper.

\subsubsection{Features}
Each pattern has 41 features, allocated to one of three categories: \textit{Basic}, \textit{Traffic}, and \textit{Content}. However, an analysis of the \textit{Mean Decrease Impurity} metric \cite{MDI_Ref} found 17 to be irrelevant. We thus considered the ez 24 features during our evaluation.

\subsubsection{Skewedness}
The skewed nature of the dataset is evident from  \Cref{kdd_data_distribution}: 98.61\% of the data belongs to either the Normal or DoS categories. As we see in later sections, this hampers the performance of classifiers on the remaining classes.

\subsubsection{Non-stationary}
The non-stationary nature of the KDD-99 dataset is clearly visible from train and test distributions in  \Cref{kdd_data_distribution}. The training set has 23\% of its data as DoS examples versus 73.9\% in the test set; Normal is 75.61\% of the train set but only 19.48\% of the test set. It has been demonstrated that such divergence negatively affects performance \cite{Cieslak_Chawla_nonstationarity_affects_performance}. Fugate and Gattiker \cite{Fugate_Gattiker_nonstationarity} investigate anomaly-detection techniques on KDD-99, accounting for this problem by segmenting the dataset and using a stationary partition for training.

\begin{table}[!t]
	\renewcommand{\arraystretch}{1.3}
	\caption{KDD CUP 1999 train and test data distribution}
	\label{kdd_data_distribution}
	\centering
	\begin{tabular}{crrrr}
		\toprule
		
		\bfseries Class & \bfseries Training Set & \bfseries Percentage & \bfseries Test Set & \bfseries Percentage
		\\
		\midrule
		Normal 	& 812,814 	& 75.611\% 	& 60,593 	& 19.481\%	\\
		DoS 	& 247,267 	& 23.002\%	& 229,853 	& 73.901\%	\\
		Probe 	& 13,860 	& 1.289\% 	& 4,166 	& 1.339\%	\\
		R2L 	& 999 		& 0.093\% 	& 16,189 	& 5.205\%	\\
		U2R 	& 52 		& 0.005\% 	& 228 		& 0.073\%	\\
		\midrule
		Total 	& 1,074,992 & 100\%  	& 311,029	& 100\%
		\\ \bottomrule
	\end{tabular}
\end{table}

\begin{table}[!t]
	\renewcommand{\arraystretch}{1.3}
	\caption{NSL-KDD train and test data distribution}
	\label{nsl_kdd_data_distribution}
	\centering
	\begin{tabular}{crrrr}
		\toprule
		
		\bfseries Class & \bfseries Training Set & \bfseries Percentage & \bfseries Test Set & \bfseries Percentage
		\\
		\midrule
		Normal 	& 67,342 & 53.458\% 	& 9,710 & 43.075\% 	\\
		DoS 	& 45,927 & 36.458\% 	& 7,457 & 33.080\% 	\\
		Probe 	& 11,656 & 9.253\% 		& 2,421 & 10.740\% 	\\
		R2L 	& 995 	 & 0.790\% 		& 2,754 & 12.217\% 	\\
		U2R 	& 52 	 & 0.041\% 		& 200 	& 0.887\% 	\\
		\midrule
		Total 	& 125,972 & 100\%  		& 22,542	& 100\%
		\\ \bottomrule
	\end{tabular}
\end{table}

\subsection{NSL-KDD}
NSL-KDD is an effort by Tavallaee et al. \cite{Tavallaee_et_al} to rectify KDD-99 and overcome its drawbacks. However, as the authors mention, the dataset is still subject to certain problems, such as its non-representation of low footprint attacks \cite{McHugh_DARPA}. 

The following aspects of NSL-KDD mark an improvement over KDD-99.

\subsubsection{Size and redundancy}
NSL-KDD has fewer data points than KDD-99, all of which are unique. It is thus less computationally expensive to use for training machine learning models.

\subsubsection{Features}

As with KDD-99, certain parameters were found unnecessary. A reduced set of 20 features found by Mean Decrease Impurity was used in this paper.

\subsubsection{Skewedness}
NSL-KDD includes an undersampling of the classes Normal, DoS and Probe. This mitigates some of the problems associated with KDD-99's skewedness; \Cref{Results_NSL_vs_KDD} compares the magnitude of improvement this makes to classifier performance.

\subsubsection{Non-Stationarity}
\Cref{nsl_kdd_data_distribution} shows the resampled distribution of data that is NSL-KDD. The target spread of the train and test sets are significantly more alike. NSL-KDD is a stationary sampling of KDD~CUP~99.

\begin{table}[!t]
	\renewcommand{\arraystretch}{1.3}
	\caption{UNSW-NB15 train and test data distribution}
	\label{nb15_data_distribution}
	\centering
	\setlength{\tabcolsep}{4pt}
	\begin{tabular}{crrrr}
		\toprule
		
		\bfseries Class & \bfseries Training Set & \bfseries Percentage & \bfseries Test Set & \bfseries Percentage
		\\
		\midrule
		Analysis 		& 2,000		& 1.141\%	& 677 		& 0.822\%	\\
		Backdoor 		& 1,746		& 0.996\%	& 583 		& 0.708\%	\\
		DoS 			& 12,264	& 6.994\%	& 4,089 	& 4.966\%	\\
		Exploits 		& 33,393	& 19.045\%	& 11,132	& 13.521\%	\\
		Fuzzers 		& 18,184	& 10.371\%	& 6,062 	& 7.363\%	\\
		Generic 		& 40,000	& 22.813\%	& 18,871	& 22.921\%	\\
		Normal 			& 56,000	& 31.938\%	& 37,000	& 44.940\%	\\
		Reconnaissance	& 10,491	& 5.983\%	& 3,496 	& 4.246\%	\\
		Shell Code 		& 1,133 	& 0.646\%	& 378 		& 0.459\%	\\
		Worms 			& 130 		& 0.074\%	& 44 		& 0.053\%	\\
		\midrule
		Total 			& 175,341 	& 100\%		& 82,332 	& 100\%
		\\ \bottomrule
	\end{tabular}
\end{table}

\subsection{UNSW-NB15} \label{NB15_desc}
The UNSW-NB15 dataset \cite{Moustafa_Slay_NB15_description} is a modern KDD-99 alternative . It is similar enough to KDD-99 in terms of its method of generation and features to be a viable substitute, but eliminates (or at least mitigates) several lacunas that make KDD-99 inconvenient for use in a contemporary NIDS.

\subsubsection{Genesis}
This dataset was simulated using the IXIA PerfectStorm tool at the ACCS (Australian Center of Cyber Security) over two days, in sessions of 16 hours and 15 hours. 45 unique IP addresses were used over 3 networks, compared to 11 IP addresses on 2 networks for KDD. Attacks were chosen from a constantly-updated CVE site, while normal behavior was not simulated. Packet-level traffic was captured via \textit{TCPdump}. A total of 2,540,044 records were generated which can be found at the ADFA website \cite{UNSW_NB15_dataset}. A much smaller partition was used for the UNSW-NB15 training and test data.

\subsubsection{Target classes}
NB15 was intended as a step up from the archaic KDD dataset, and it incorporates 10 targets: one Normal, and 9 anomalous, namely: Fuzzers, Analysis, Backdoors, DoS, Exploits, Generic, Reconnaissance, Shell Code and Worms.

The greater number of target classes (10 in NB15, vs. 5 in KDD) and a higher \textit{Null Error Rate} (55.06\% in NB15 vs. 26.1\% in KDD) consequently has a mitigating effect on the overall classifier accuracy. For this reason, in  \Cref{results} we also consider the performance on binary UNSW-NB15, viz the dataset with only \textit{Normal} and an aggregated \textit{Attack} classes. We compare this to likewise binary versions of other datasets.

\subsubsection{Size and redundancy}
The train set contains 175,341 data points and test set 82,332. While these figures are a fraction of KDD-99, we found the size sufficient to train high-variance classifiers for intrusion detection. There are no redundant data points.

\subsubsection{Features}
49 Features were extracted using the Argus and Bro-IDS tools and categorized into five groups: \textit{Flow}, \textit{Basic}, \textit{Content}, \textit{Time}, and \textit{Additionally Generated}. Once again, the Mean Decrease Impurity metric filters out irrelevant features. The reduced feature-set of size 30 is used.

\subsubsection{Skewedness}
\begin{figure}[!h]
	\centerline{\includegraphics[scale=0.06]{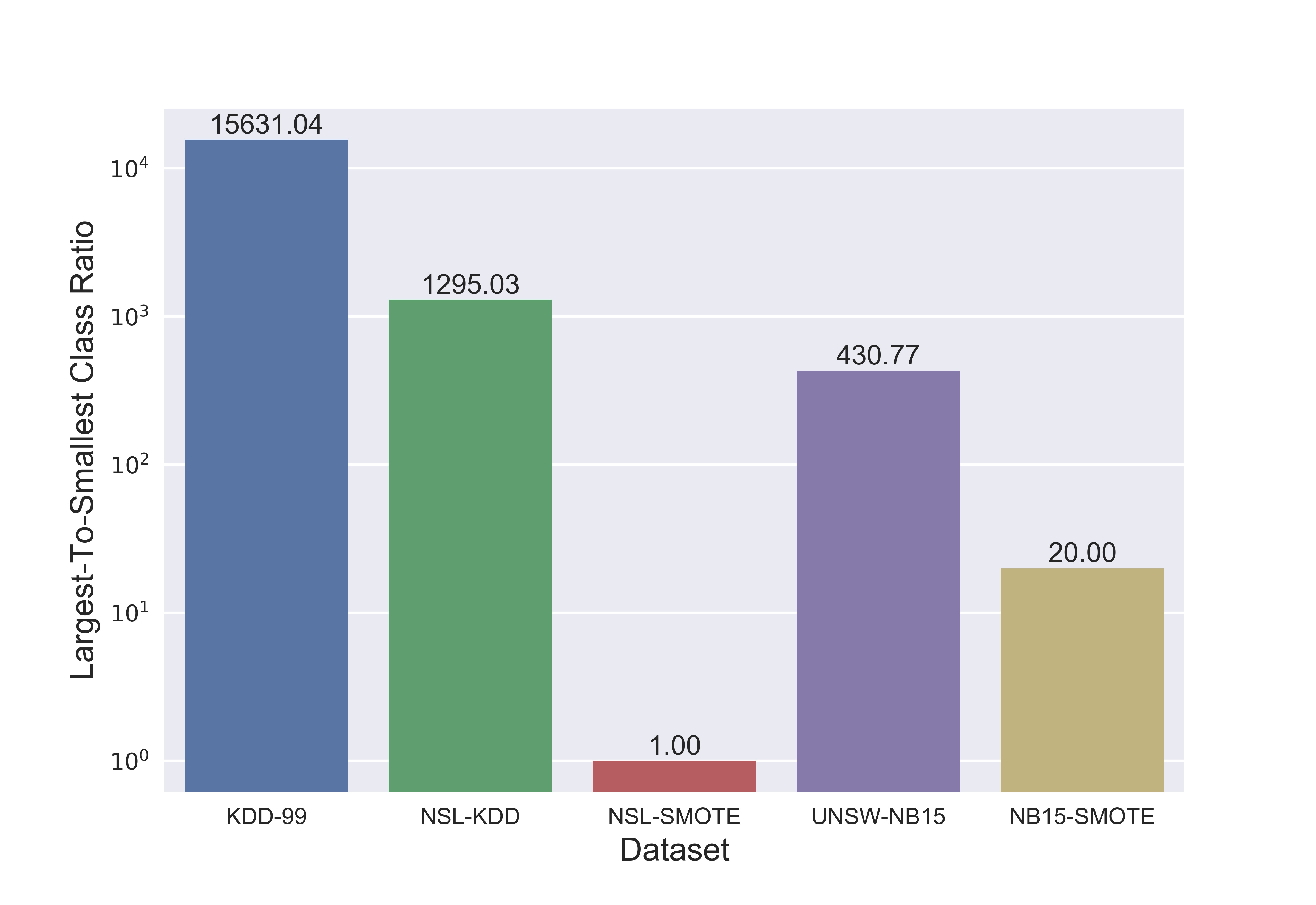}}
	\caption{Largest-To-Smallest class ratio across datasets; KDD-99 is by far the most skewed. UNSW-NB15 is measurably more uniform, especially for the larger classes. SMOTE oversampling balances the datasets significantly.}
	\label{fig: largest_to_smallest_graph}
\end{figure}

We highlight the uniformity of UNSW-NB15 compared to the traditional datasets by observing the largest-to-smallest target ratio in  \Cref{fig: largest_to_smallest_graph}. KDD-99 is the most imbalanced. NSL-KDD attempts to alleviate this problem. At a glance, UNSW-NB15's skewedness is noticeably lower, a fact which is corroborated by Moustafa and Slay's statistical analysis \cite{Moustafa_Slay_NB15_statistical}.

\subsubsection{Non-Stationarity} 
Data stationarity is maintained between the training and test sets in NB15; as seen in  \Cref{nb15_data_distribution}, both have similar distributions.

\section{Methodology} \label{methodology}

\begin{figure*}[!h]
	\centering \includegraphics[scale=0.45]{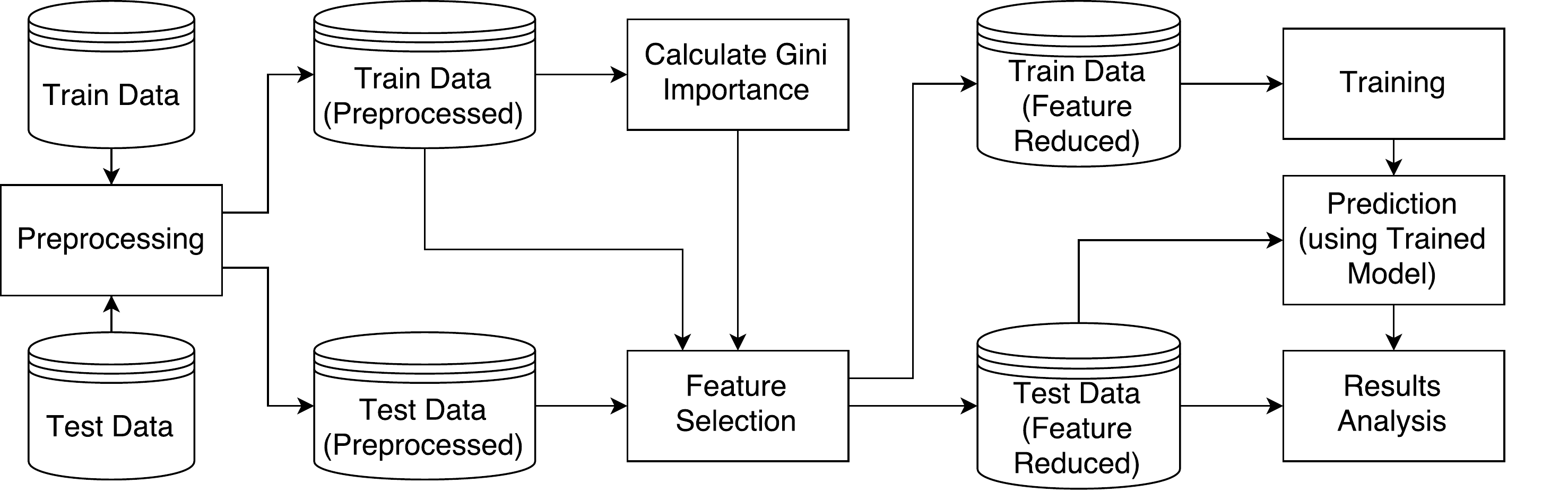}
	\caption{Machine Learning Pipeline}
	\label{fig: pipeline}
\end{figure*}

In this section various machine learning techniques have been utilized to measure classification performance on the datasets and highlight their characteristics.

We follow the Machine Learning pipeline outlined in  \Cref{fig: pipeline}. The \textit{Preprocessing}, \textit{Feature Selection}, \textit{Training} and \textit{Prediction} stages are the most pertinent. As part of \textit{Preprocessing}, we construct the \textit{NSL-SMOTE} and \textit{NB15-SMOTE} datasets using the SMOTE oversampling process and then random undersampling. The \textit{Training} stage yields a \textit{Trained Model}, which is assessed on the feature-reduced \textit{Test Dataset}. We apply the pipeline to each of the datasets KDD-99, NSL-KDD, NSL-SMOTE, UNSW-NB15 and NB15-SMOTE. Results are explored in  \Cref{results}.

\subsection{Preprocessing and Feature Selection}
\begin{enumerate}
	\item All three datasets have features with string values. We map each unique string value to an integer.
	
	\item KDD-99 faces high redundancy in both its training and test set. The redundant examples were filtered for computational convenience and to deter classifiers from skewing towards the highly repeated classes. 
	
	\item To remove all irrelevant features we first rank these features using the Gini Impurity Index \cite{MDI_Ref}. 
	
	\quad We select predictors that have non-zero MDI. For the KDD-99, only 24 of the 41 features were found to have non-zero MDI. NSL-KDD was similarly reduced to 20, and UNSW-NB15 dropped from 49 to 30.
	
	\item We use \textit{mean standardization} for feature scaling to ensure all predictor values lie in a similar range. Mean standardization centers each feature's data vector around the mean of the distribution and scales by the standard deviation. 
\end{enumerate}

\subsection{Oversampling and Undersampling}
Yap B.W. et al. \cite{yap} suggests methods to achieve uniform distribution of pattern responses; we have elected to use SMOTE and random undersampling. The Python library \textit{imbalance-learn} (version 0.2.1) \cite{imbalance-learn} was used for this purpose.

SMOTE \cite{SMOTE} is an oversampling technique used to boost the counts of imbalanced targets. It is applied to NSL-KDD and UNSW-NB15; the resulting training sets are referred to as \textit{NSL-SMOTE} and \textit{NB15-SMOTE} respectively. The test sets used for Prediction remain unchanged. 

As SMOTE followed by undersampling on KDD-99 would result in a dataset that is functionally identical to NSL-SMOTE, we refrain from doing so.

\begin{enumerate}
	\item SMOTE (Synthetic Minority Oversampling Technique): We alter the NSL-KDD training set to establish a ratio of 0.015 for U2R (minority) to Normal (majority), resulting in slightly more than a thousand U2R examples. UNSW-NB15's Analysis and Backdoor are oversampled to 14,000 examples each (ratio of 0.25 to the Normal class), and Shell Code and Worms to 2,800 each.

	\item Random undersampling: To balance each class of NSL-SMOTE to 995 examples, random undersampling was used for the DoS, Normal, Probe are U2R targets. 995 was chosen as it is the number of data points in the R2L class, which is now the minority class after oversampling U2R. Random undersampling was not used during the construction of NB15-SMOTE.
	
\end{enumerate}

\subsection{Training}
As it is not our objective to explore complex learning methodologies on the datasets discussed, we utilize a standard package for the various classifiers discussed in this paper: Python's \textit{scikit-learn} library (version 0.18.1) \cite{scikit-learn}. We have found that classifiers trained using this package converge to a local minima on a suitable timescale when dealing with hundreds of thousands of data points.

Six machine learning models were implemented to establish benchmarks. For each classifier, hyperparameters were tuned using K-fold Cross Validation on each of the datasets with \mbox{$K=5$}. Randomized Grid Search was used to obtain the respective hyperparameter values.

\begin{enumerate}
	\item Naive Bayes: we use the default of Gaussian Naive Bayes.
	\item Support Vector Machine: a Radial Basis Function kernel was used.
	\item Decision Tree: both Entropy and Gini index were used as splitting criterion; in each particular case, the one yielding the highest cross-validation accuracy was used. The depth was selected up to a maximum of 30.
	\item Random Forest: criterion similar to Decision Tree were used. The number of trees was cross-validated on up to a maximum of 100.
	\item Neural Network: architectures with between 1-3 hidden layers were evaluated, with 20-100 neurons per layer.
	\item K-Means using Majority Vote: we first apply K-Means to group the dataset into 300-400 clusters. 
	
	During Prediction, each test pattern is assigned to the cluster with the closest centroid in Euclidean distance. The target is selected as the majority vote of the targets in that cluster (with random assignment serving as a tie-breaker).
	
\end{enumerate}

\subsection{Analysis Metrics}
When analyzing the performance of classifiers on each dataset, prevalent metrics we consider are Precision, Recall and F1-Score. Values for each metric are calculated from the confusion matrix of predictions.

\begin{enumerate}
	\item F1-Score: It is calculated as the harmonic mean of precision and recall. Gives a single measure of comparison. Higher is better.
	\begin{equation}
	\label{f1-score}
	F\textit{1}\textnormal{-}Score = \frac{2PR}{P+R}
	\end{equation}
	Where \textit{P} is Precision and \textit{R} is Recall
	
	\item Weighted F1-Score: the average of F1-Scores over all classes of the dataset, each weighted by its Support. 
	\begin{equation}
	\label{avg-f1-score}
	Weighted~F\textit{1}\textnormal{-}Score = 
	\frac{\sum\limits_{i=1}^{K} Support_i \cdot F\textit{1}_i }{Total}
	\end{equation}
	where $F\textit{1}_i$ is the F1-Score predicted for the $i^{th}$ target class.
	
	\item Null Error Rate (NER): a measure of the misclassification error inherent in the test dataset if our classifier naively assigned every test pattern to the majority class.
	\begin{equation}
	\label{null_error_rate}
	NER = 1 - \frac{Support_m}{Total}
	\end{equation}
	where $Support_m$ is the number of examples in the majority class.
\end{enumerate}			

We discount the pertinence of accuracy in the current scenario. Classification accuracy is an insufficient metric when assessing imbalanced datasets, as models can exhibit high accuracy scores even while misclassifying all but a few examples in minority classes \cite{accuracy_paradox}. KDD's skewedness and its Null Error Rate (26.1\%) versus that of a perfectly balanced dataset with as many targets (80\%), makes it susceptible to this misconception. 

However, despite common knowledge of this phenomenon, a majority of works found during the literature study use accuracy as their primary means of evaluation. In the Network Intrusion Detection context, where even a meager number of false negatives can be catastrophic, accuracy paints an overly optimistic view of system security. 

To circumvent the accuracy paradox, \textit{F1-score} is used to measure performance. This is a combined metric, the harmonic mean of Precision and Recall. We present F1-Scores for each model on a class-wise basis.

When we require a single variable to depict the overall classifier performance we use Weighted F1-Score. Our argument is that a skewed summation of class-wise F1-Scores would avoid the paradox more effectively than simple accuracy. In the following section, we analyze dataset-wise as well as class-wise performance using both these metrics.

\section{Results} \label{results}

\begin{figure}[!h]
	\centerline{\includegraphics[scale=0.055]{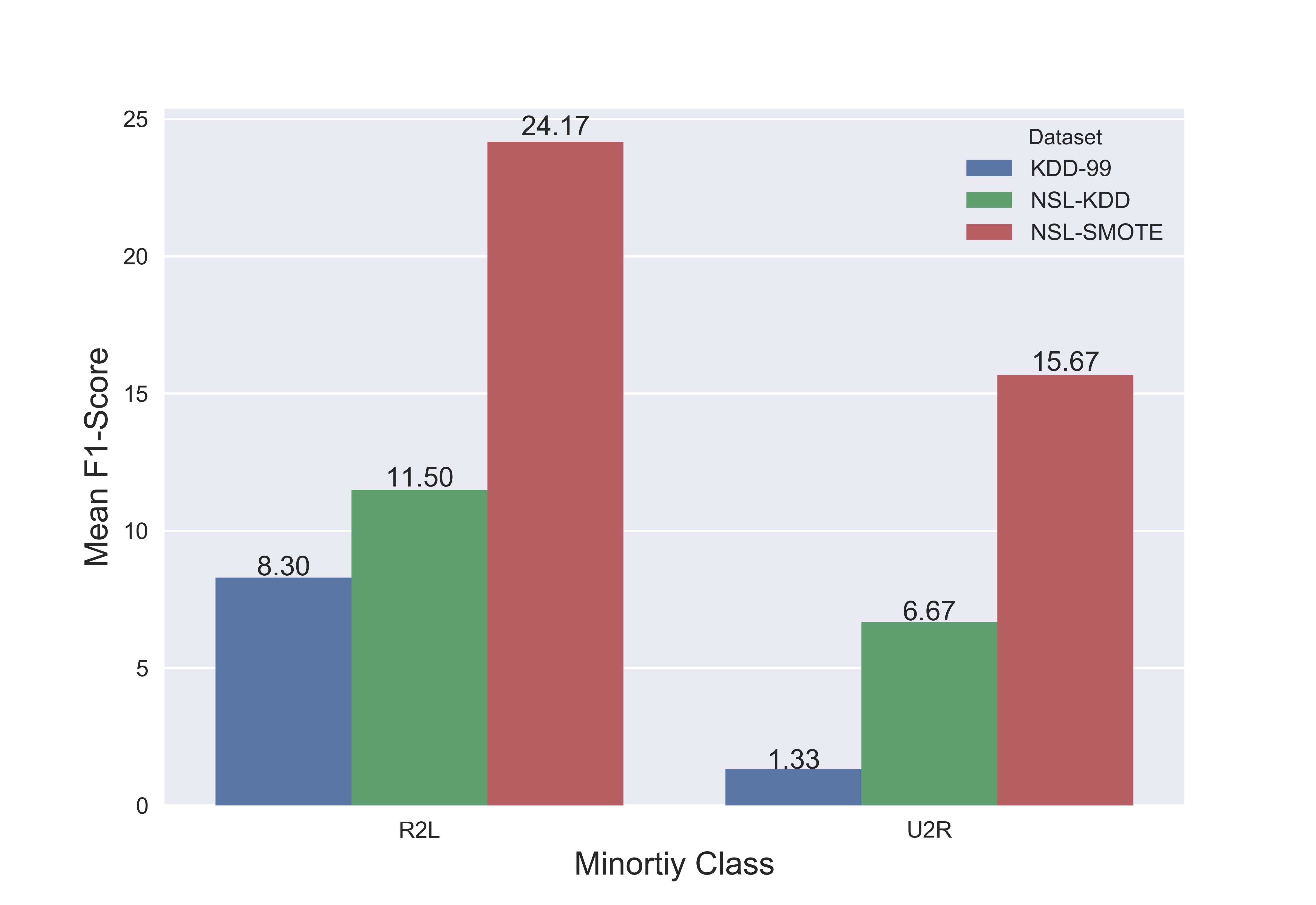}}
	\caption{Mean F1-Scores of all models on minority classes R2L and U2R.}
	\label{fig: minory_classes_f1_scores}
\end{figure}	

\begin{table}[!t]
	\renewcommand{\arraystretch}{1}
	\caption{A comparison of class-wise and Weighted F1-Scores for KDD-99, NSL-KDD, and NSL-SMOTE.}
	\label{kdd_vs_nsl_vs_nsl_smote}
	\centering
	
	\begin{tabular}{
			@{\hskip1pt}c@{\hskip1pt}
			c
			c@{\hskip1pt}
			@{\hskip1pt}c@{\hskip1pt}
			@{\hskip1pt}c@{\hskip1pt}
			@{\hskip1pt}c@{\hskip1pt}
			@{\hskip1pt}c
			b{1cm}} 
		\toprule
		
		\multirow{2}{*}{\bfseries Model} 
		& \multirow{2}{*}{\bfseries Dataset} 
		& \multicolumn{5}{c}{\bfseries F1-Score} 
		& \multirow{2}{*}{\bfseries Weighted} \\ \cmidrule{3-7} 
		& & \bfseries DoS & \bfseries Normal & \bfseries Probe & \bfseries R2L & \bfseries U2R & \bfseries \mbox{F1-Score}
		
		\\ \midrule
		
		\multirow{3}{*}{Naive Bayes}
		& KDD-99 & 40 & 37 & 75 & 18 & 1 			& 39 \\ 
		& NSL-KDD & 63 & 73 & 7 & 29 & 4 			& 57 \\ 
		& NSL-SMOTE & 63 & 73 & 7 & 28 & 5 			& 57 \\
		\midrule
		
		\multirow{3}{*}{SVM}
		& KDD-99 & 99 & 83 & 74 & 6 & 0 			& 90 \\ 
		& NSL-KDD & 84 & 80 & 72 & 31 & 12 			& 74 \\ 
		& NSL-SMOTE& 78 & 79 & 69 & 24 & 14 		& 71 \\
		\midrule
		
		\multirow{3}{*}{Decision Tree}
		& KDD-99 & 98 & 84 & 71 & 8 & 4 			& 90 \\ 
		& NSL-KDD & 84 & 77 & 66 & 3 & 12 			& 68 \\ 
		& NSL-SMOTE& 79 & 79 & 71 & 23 & 12 		& 71 \\
		\midrule
		
		\multirow{3}{*}{Random Forest}
		& KDD-99 & 97 & 80 & 83 & 6 & 2				& 89 \\ 
		& NSL-KDD & 87 & 79 & 70 & 0 & 4			& 70 \\ 
		& NSL-SMOTE& 85 & 81 & 61 & 14 & 28			& 72 \\
		\midrule
		
		\multirow{3}{*}{\mbox{Neural} Network}
		& KDD-99 & 98 & 84 & 83 & 8 & 2				& 91 \\ 
		& NSL-KDD & 85 & 79 & 74 & 3 & 5			& 71 \\ 
		& NSL-SMOTE& 75 & 80 & 68 & 31 & 18			& 70 \\
		\midrule
		
		\multirow{3}{*}{\mbox{K-Means}}
		& KDD-99 & 98 & 84 & 84 & 4 & 0				& 90 \\ 
		& NSL-KDD & 84 & 78 & 52 & 3 & 3			& 68 \\ 
		& NSL-SMOTE& 66 & 79 & 57 & 25 & 14			& 65 \\
		\bottomrule
	\end{tabular}
\end{table}

\subsection{KDD-99 vs. NSL-KDD} \label{Results_NSL_vs_KDD}

To demonstrate how imbalance hampers the performance of classifiers on the smallest classes (U2R and R2L), we have analyzed NSL-KDD before and after applying SMOTE and random undersampling.  \Cref{fig: minory_classes_f1_scores} contrasts the results with KDD-99. For the R2L class we observe a significant increase in the F1-Score from KDD-99 to NSL-SMOTE (18 to 28, 6 to 24, 8 to 23, 8 to 31, 4 to 23). Significant gains are also achieved from NSL-KDD to NSL-SMOTE (1 to 37 for Decision Tree, 7 to 32 for Neural Network, 0 to 25 for K-Means). A full performance breakdown is detailed in  \Cref{kdd_vs_nsl_vs_nsl_smote}.

It is interesting to note that there is a persistent decrease in F1-Score for the DoS and Probe classes from KDD-99 to NSL-SMOTE. This can be attributed to the inflated counts of DoS patterns in KDD-99; with the undersampling, the classifiers trained on NSL-SMOTE are forced to concentrate equally on all classes, thus underfitting the DoS class compared to KDD. NSL-KDD, which constitutes frequently misclassified records in KDD and is more balanced, has F1-Scores similar to NSL-SMOTE for the DoS class.

\begin{table}[!t]
	\renewcommand{\arraystretch}{1.3}
	\caption{Class-wise and Weighted F1-Scores for UNSW-NB15}
	\label{unsw_nb15_results}
	\centering
	
	\begin{tabular}{
			m{0.22\linewidth}   @{\hskip2pt}m{0.10\linewidth}@{\hskip2pt}    @{\hskip2pt}m{0.08\linewidth}@{\hskip2pt}    @{\hskip2pt}m{0.12\linewidth}@{\hskip2pt}    @{\hskip2pt}m{0.12\linewidth}@{\hskip2pt}    @{\hskip2pt}m{0.12\linewidth}@{\hskip2pt}    @{\hskip2pt}m{0.12\linewidth}@{\hskip2pt}  }
		\toprule
		\bfseries Class & \bfseries Naive	Bayes & \bfseries SVM 	& \bfseries Decision Tree	& \bfseries Random Forest	& \bfseries \mbox{Neural} \mbox{Network} 	& \bfseries \mbox{K-Means} \\ \midrule
		Analysis 										& 2		&	0		&	5		&	0		&	2		&	0			\\
		Backdoor 										& 6		&	0		&	5		&	7		&	0		&	0			\\
		DoS													& 1		&	6		&	19	&	14	&	10	&	5			\\
		Exploits 										& 42	&	64	&	64	&	70	&	64	&	60		\\
		Fuzzers 										& 21	&	33	&	35	&	38	&	36	&	32		\\
		Generic 										& 73	&	98	&	98	&	98	&	97	&	96		\\
		Normal 											& 56	&	79	&	82	&	84	&	74	&	74		\\
		Reconnaissance 							& 14	&	64	&	85	&	86	&	56	&	47		\\
		Shell Code 									& 4		&	6		&	46	&	45	&	0		&	0			\\
		Worms												& 3		&	13	&	56	&	25	&	4		&	0			\\ \midrule
		
		\bfseries Weighted \mbox{F1-Score} &	50	&	72	&	76	&	77	&	70	& 68		\\ \bottomrule
	\end{tabular}
\end{table}

\begin{table}[!t]
	\renewcommand{\arraystretch}{1.3}
	\caption{Class-wise and Weighted F1-Scores for NB15-SMOTE}
	\label{nb15_smote_results}
	\centering
	
	\begin{tabular}{
			m{0.22\linewidth}   @{\hskip2pt}m{0.10\linewidth}@{\hskip2pt}    @{\hskip2pt}m{0.08\linewidth}@{\hskip2pt}   @{\hskip2pt}m{0.12\linewidth}@{\hskip2pt}    @{\hskip2pt}m{0.12\linewidth}@{\hskip2pt}    @{\hskip2pt}m{0.12\linewidth}@{\hskip2pt}    @{\hskip2pt}m{0.12\linewidth}@{\hskip2pt}  }
		\toprule
		\bfseries Class & \bfseries Naive	Bayes & \bfseries SVM 	& \bfseries Decision Tree	& \bfseries Random Forest	& \bfseries \mbox{Neural} \mbox{Network} 	& \bfseries \mbox{K-Means} \\ \midrule
		Analysis 											& 1		& 2 	&	6		&	13	&	4		&	4			\\
		Backdoor 											& 6		& 16 	&	4		&	7		&	2		&	6			\\
		DoS 													& 1		& 15 	&	22	&	21	&	11	&	2			\\
		Exploits 											& 43	& 62 	&	62	&	71	&	63	&	59		\\
		Fuzzers 											& 21	& 31 	&	32	&	39	&	33	&	31		\\
		Generic 											& 73	& 94 	&	98	&	98	&	97	&	95		\\
		Normal 												& 55	& 81 	&	83	&	84	&	81	&	74		\\
		Reconnaissance 								& 4		& 62 	&	84	&	87	&	53	&	49		\\
		Shell Code 										& 4		& 14 	&	34	&	41	&	25	&	16		\\
		Worms 												& 3		& 0 	&	35	&	65	&	21	&	5			\\ \midrule
		
		\bfseries Weighted \mbox{F1-Score} 	&	49	&	72	&	75	&	78	&	72	& 68		\\ \bottomrule
	\end{tabular}
\end{table}

\begin{table}[!t]
	\renewcommand{\arraystretch}{1.3}
	\caption{Comparison of Weighted F1-scores over all test sets (Binary)}
	\label{comparison_of_f1_scores_for_all_datasets_binary}
	\centering
	\begin{tabular}{cccc}
		\toprule
		\bfseries Model & \bfseries KDD-99 & \bfseries NSL-KDD & \bfseries UNSW-NB15\\
		\midrule
		Naive Bayes & 86 & 78 & 76\\
		SVM & 94 & 82 & 82\\
		Decision Tree & 94 & 81 & 86.5\\
		Random Forest & 94 & 81 & 88.5\\
		Neural Networks & 94 & 81 & 84.88\\
		K-Means & 93 & 83 & 83\\
		\bottomrule
	\end{tabular}
\end{table}

\subsection{UNSW-NB15} \label{Results_NB15}
We argue that one of the major hurdles faced by researchers in the domain of Network Intrusion Detection is the paucity of results available for datasets other than KDD-99. UNSW-NB15 was found to be a modern substitute which currently lacks exposure, and hence we analyze the aforementioned models with respect to this dataset. Additionally, we perform SMOTE oversampling to highlight the effect of class imbalance even in UNSW-NB15. Unlike other works \cite{Moustafa_Slay_NB15_statistical}, we consider primarily the Weighted F1-Score, as accuracy is misleading for such data distributions.

From  \Cref{unsw_nb15_results}, UNSW-NB15 is seen to exhibit high performance on the Exploits, Generic, and Normal classes, which together consist over 73\% of training data. The four responses with the fewest patterns - Analysis, Backdoor, Shell Code and Worms - make up just 2.857\% of the data. Not surprisingly, learning models do not perform well on these targets. The exceptions seem to be Decision Tree and Random Forest models. DoS performance remains weak as well, despite a fair number of training patterns. 

The application of SMOTE increases the support of Shell Code and Worms to 2,800, and Analysis and Backdoor to 14,000. These classes together now constitute 16.48\% of the data. We see an improvement; the mean F1-Score of Shell Code, for example, increases from 16.8 to 22.3. The total mean increase for these four classes is 17.5 points. The largest increase is Random Forest (from 25 to 65 in Worms). 

As the datasets have different responses a direct comparison is infeasible. Instead, we analyze classification on binarized versions of KDD-99, NSL-KDD and UNSW-NB15, seen in  \Cref{comparison_of_f1_scores_for_all_datasets_binary}. The Weighted F1-Score obtained for KDD-99 is once again high, but as we saw earlier in  \Cref{kdd_vs_nsl_vs_nsl_smote}, the individual F1-Scores on minority classes in KDD-99 is anemic. However, binarizing the data effectively hides this discrepancy. As for the others, it is evident that UNSW-NB15 equals or betters NSL-KDD on almost all learning models implemented.

\section{Discussion} \label{discussion}
The previous section details the results of the investigation. It is seen that KDD's highly oblique distribution of patterns and non-stationarity between train and test sets, evident in \Cref{kdd_data_distribution}, leads us to question the effectiveness of techniques optimized on KDD-99. Observing  \Cref{kdd_vs_nsl_vs_nsl_smote}, the minority F1-Scores were found to be unsatisfactory: best scores for R2L were 18 and 31 on KDD-99 and NSL-KDD respectively, and 4 and 12 for U2R. In contrast, the DoS F1-Score neared 98 on KDD-99 and 85 on NSL-KDD for most cases. Such discrepancy in performance was obscured when considering the accuracy and even Weighted F1-Scores. The problem is exacerbated by binarization. We infer that machine learning techniques optimized on KDD-99 and NSL-KDD may be exposed to minority class attacks while claiming a higher efficacy.

An imperative requirement is target uniformity. If a dataset is balanced, the Null Error Rate grows in proportion to the number of target classes, and multiclass classification is subsequently more difficult. The converse occurs when targets are imbalanced. Guo et al. explain this phenomenon and suggest measures to counter it, including forgoing accuracy for other metrics and oversampling smaller classes \cite{class_imbalance_problem}. As a result, the SMOTE technique \cite{SMOTE} was used to oversample the U2R class and construct \textit{NSL-SMOTE}, a perfectly balanced dataset (refer  \Cref{methodology}). As seen in  \Cref{fig: minory_classes_f1_scores}, this improved performance on the minority classes: the mean F1-Score increased from NSL-KDD to NSL-SMOTE by factors of 2.35 for U2R and 2.1 for R2L. We thus verify that dataset asymmetry plays a role in the performance of KDD-99's minority classes. However, absolute scores remained marginal, suggesting that the lack of natural diversity in the underlying data (52 for U2R) led to oversampling of noisy and borderline examples \cite{Saez_SMOTE_noisy_borderline}. Additional legitimate attack patterns might benefit detection.

In the latter half of our study we benchmarked \mbox{UNSW-NB15} - a contemporary network intrusion dataset. \mbox{UNSW-NB15} is less skewed than \mbox{KDD-99} or \mbox{NSL-KDD} (refer  \Cref{fig: largest_to_smallest_graph}). However, with a largest-to-smallest class ratio of 430, it still qualifies as imbalanced, and as a result minority F1-Scores suffer (\Cref{unsw_nb15_results}). SMOTE oversampling improves the situation, especially for Analysis and Backdoor (\Cref{nb15_smote_results}). Observing the confusion matrix (not included) reveals that Analysis, Backdoor and DoS are often misclassified as Exploits, regardless of the oversampling ratio selected; selective oversampling may lead to better results.

The best performer was found to be Random Forest \cite{RandomForest_Breiman}. In this model, each tree is trained and becomes an ‘expert’ on a randomly-split subset of data; consequently, trees trained on the oversampled minority classes are able to better classify them. Increasing the support thus allows more such trees to contribute to the majority vote. These results suggest that bagging and boosting methods such as AdaBoost and SMOTEBoost \cite{SMOTEBoost} may positively affect the performance of minority targets. On the other hand, Gaussian Naive Bayes was ineffectual in all scenarios explored. Ihe inflexibility of the model on features with dependencies \cite{Rish_naive_bayes_properties} seems to be the contributing factor.

Binarizing the classes eliminates the problem of imbalance. It was found that NB15 equals or betters the Weighted F1-Score of NSL-KDD; as seen in  \Cref{comparison_of_f1_scores_for_all_datasets_binary}, maximum scores were 83 \& 88.5, using NSL-KDD \& UNSW-NB15 respectively. This leads us to reason that UNSW-NB15 would train an adequate binary anomaly detector, often used as for preliminary filtering in a multistage NIDS.

In summary the results strongly indicate that UNSW-NB15 can satisfactorily substitute the archaic KDD~CUP~99 dataset and even NSL-KDD when used to train machine learning anomaly-based NIDSs. We thus encourage its use in forthcoming research.

\section{Conclusion} \label{conclusion}
Despite common knowledge of its flaws \cite{McHugh_DARPA}, KDD~CUP~99 and its parent DARPA98 have remained some of the most widely used datasets in the annals of Anomaly-based Network Intrusion Detection \cite{Tsai_intrusion_detection_machine_learning_review, Shamshirband_et_al_wireless}, possibly due to a lack of substitute datasets. 

In this paper, we have evaluated UNSW-NB15, a dataset with 10 modern attack classes and less skewed target distribution. F1 performance has been benchmarked on models commonly trained on KDD-99 and NSL-KDD, to facilitate UNSW-NB15's adoption in future research. Moreover, we have demonstrated how oversampling overcomes the anemic performance of classifiers trained on NSL-KDD and UNSW-NB15.

The results of this paper leave sufficient scope to optimize performance by using alternative techniques across the machine learning pipeline in \Cref{fig: pipeline}. 

For example: \mbox{(1) SMOTE} or one of its popular variants \cite{Han_Borderline_SMOTE} could be used to balance the smaller classes in UNSW-NB15. \mbox{(2) Random} undersampling as used in this paper leads to several majority examples being ignored. To overcome this, undersampling methods like EasyEnsemble and BalanceCascade \cite{Liu_undersampling} may be used instead. \mbox{(3) Ensemble} methods other than RandomForest have not been explored and have the potential to vastly improve the F1 performance on minority classes \cite{galar_ensembles_for_imbalanced_classes}. \mbox{(4) Clustering} techniques such as those proposed by Wang et al. \cite{Wang_et_al}. could be used for preprocessing or outlier detection. Unsupervised learning may also be used for classification, such as the hierarchical Self-Organizing Maps (SOM) architectures \cite{Kayacik_SOM}. The authors find that a 6-feature, two-layer architecture (using both Kohonen and other SOMs) optimized on KDD-99. \mbox{(5) Hybrid} approaches \cite{Tsai_intrusion_detection_machine_learning_review} taken to analyze KDD-99 -- especially involving high-variance models such as SVMs and Neural Networks -- may be applicable to UNSW-NB15.

Finally, only a small portion of the ACCS labs data \cite{UNSW_NB15_dataset} was used to configure UNSW-NB15; the total dataset of over 2.5 million records remains largely unexplored.

\bibliographystyle{IEEEtran}

\begin{thebibliography}{10}
	\providecommand{\url}[1]{#1}
	\csname url@samestyle\endcsname
	\providecommand{\newblock}{\relax}
	\providecommand{\bibinfo}[2]{#2}
	\providecommand{\BIBentrySTDinterwordspacing}{\spaceskip=0pt\relax}
	\providecommand{\BIBentryALTinterwordstretchfactor}{4}
	\providecommand{\BIBentryALTinterwordspacing}{\spaceskip=\fontdimen2\font plus
		\BIBentryALTinterwordstretchfactor\fontdimen3\font minus
		\fontdimen4\font\relax}
	\providecommand{\BIBforeignlanguage}[2]{{%
			\expandafter\ifx\csname l@#1\endcsname\relax
			\typeout{** WARNING: IEEEtran.bst: No hyphenation pattern has been}%
			\typeout{** loaded for the language `#1'. Using the pattern for}%
			\typeout{** the default language instead.}%
			\else
			\language=\csname l@#1\endcsname
			\fi
			#2}}
	\providecommand{\BIBdecl}{\relax}
	\BIBdecl
	
	\bibitem{Patcha_et_al_overview_of_anomaly_detection_techniques_solutions_trends}
	A.~Patcha and J.-M. Park, ``{An overview of anomaly detection techniques:
		Existing solutions and latest technological trends},'' \emph{Computer
		Networks}, vol.~51, no.~12, pp. 3448--3470, 2007.
	
	\bibitem{Ilgun_et_al}
	K.~Ilgun, R.~A. Kemmerer, and P.~A. Porras, ``{State transition analysis: A
		rule-based intrusion detection approach},'' \emph{IEEE Transactions on
		Software Engineering}, vol.~21, no.~3, pp. 181--199, 1995.
	
	\bibitem{Sekar_et_al_Specification_based_anomaly_detection_a_new_approach}
	R.~Sekar, A.~Gupta, J.~Frullo, T.~Shanbhag, A.~Tiwari, H.~Yang, and S.~Zhou,
	``{Specification-based anomaly detection: A new approach for detecting
		network intrusions},'' in \emph{Proceedings of the ACM Conference on Computer
		and Communications Security}, 2002, pp. 265--274.
	
	\bibitem{Teodoro_anomaly_based_nids_techniques_systems_challenges}
	P.~Garcia-Teodoro, J.~Diaz-Verdejo, G.~Macia-Fernandez, and E.~Vazquez,
	``{Anomaly-based network intrusion detection: Techniques, systems and
		challenges},'' \emph{Computers {\&} Security, Elsevier}, vol.~28, no.~1, pp.
	18--28, 2009.
	
	\bibitem{Outside_The_Closed_World}
	R.~Sommer and V.~Paxson, ``{Outside the closed world: On using machine learning
		for network intrusion detection},'' in \emph{Symposium on Security and
		Privacy (SP), 2010 IEEE}, 2010, pp. 305--316.
	
	\bibitem{KDD_99_dataset}
	``{KDD Cup 1999 Data},''
	\url{http://kdd.ics.uci.edu/databases/kddcup99/kddcup99.html}, 1999,
	retrieved December 15, 2016.
	
	\bibitem{Shamshirband_et_al_wireless}
	S.~Shamshirband, N.~B. Anuar, M.~L.~M. Kiah, and A.~Patel, ``{An appraisal and
		design of a multi-agent system based cooperative wireless intrusion detection
		computational intelligence technique},'' \emph{Engineering Applications of
		Artificial Intelligence}, vol.~26, no.~9, pp. 2105--2127, Oct. 2013.
	
	\bibitem{Tavallaee_et_al}
	M.~Tavallaee, E.~Bagheri, W.~Lu, and A.~A. Ghorbani, ``{A detailed analysis of
		the KDD CUP 99 data set},'' in \emph{IEEE Symposium on Computational
		Intelligence for Security and Defense Applications, CISDA 2009}, 2009.
	
	\bibitem{Stolfo_et_al_KDD_creation}
	S.~J. Stolfo, W.~Fan, W.~Lee, A.~Prodromidis, and P.~K. Chan, ``{Cost-based
		modeling for fraud and intrusion detection: Results from the JAM project},''
	in \emph{Proceedings - DARPA Information Survivability Conference and
		Exposition, DISCEX 2000}, vol.~2, 2000, pp. 130--144.
	
	\bibitem{McHugh_DARPA}
	J.~McHugh, ``{Testing Intrusion detection systems: a critique of the 1998 and
		1999 DARPA intrusion detection system evaluations as performed by Lincoln
		Laboratory},'' \emph{ACM Transactions on Information and System Security},
	vol.~3, no.~4, pp. 262--294, 2000.
	
	\bibitem{KDD_DARPA_attack_description_and_schedule}
	``{DARPA98 attack description and schedule},''
	\url{https://www.ll.mit.edu/ideval/docs/attacks.html}, 1998, retrieved
	December 15, 2016.
	
	\bibitem{MDI_Ref}
	L.~Breiman, J.~H. Friedman, R.~A. Olshen, and C.~J. Stone,
	\emph{{Classification and Regression Trees}}, 1984.
	
	\bibitem{Cieslak_Chawla_nonstationarity_affects_performance}
	D.~A. Cieslak and N.~V. Chawla, ``{A framework for monitoring classifiers'
		performance: when and why failure occurs?}'' \emph{Knowledge and Information
		Systems}, vol.~18, no.~1, pp. 83--108, 2009.
	
	\bibitem{Fugate_Gattiker_nonstationarity}
	M.~Fugate and J.~R. Gattiker, ``{Anomaly detection enhanced classification in
		computer intrusion detection},'' in \emph{Pattern Recognition with Support
		Vector Machines}, vol. 2388, 2002, pp. 186--197.
	
	\bibitem{Moustafa_Slay_NB15_description}
	N.~Moustafa and J.~Slay, ``{UNSW-NB15: a comprehensive data set for network
		intrusion detection systems (UNSW-NB15 network data set)},'' in \emph{2015
		Military Communications and Information Systems Conference (MilCIS)}, 2015,
	pp. 1--6.
	
	\bibitem{UNSW_NB15_dataset}
	``{UNSW-NB15 dataset},''
	\url{http://www.unsw.adfa.edu.au/australian-centre-for-cyber-security/cybersecurity/ADFA-NB15-Datasets},
	2015, retrieved December 15, 2016.
	
	\bibitem{Moustafa_Slay_NB15_statistical}
	N.~Moustafa and J.~Slay, ``{The evaluation of Network Anomaly Detection
		Systems: Statistical analysis of the UNSW-NB15 data set and the comparison
		with the KDD99 data set},'' \emph{Information Security Journal: A Global
		Perspective}, vol.~25, no. 1-3, pp. 18--31, Apr. 2016.
	
	\bibitem{yap}
	B.~W. Yap, K.~A. Rani, H.~A. {Abd Rahman}, S.~F., Z.~Khairudin, and N.~N.
	Abdullah, ``{An application of oversampling, undersampling, bagging and
		boosting in handling imbalanced datasets},'' in \emph{Lecture Notes in
		Electrical Engineering}, vol. 285 LNEE, 2014, pp. 13--22.
	
	\bibitem{imbalance-learn}
	G.~Lema{{\^i}}tre, F.~Nogueira, and C.~K. Aridas, ``Imbalanced-learn: A python
	toolbox to tackle the curse of imbalanced datasets in machine learning,''
	\emph{Journal of Machine Learning Research}, vol.~18, no.~17, pp. 1--5, 2017.
	
	\bibitem{SMOTE}
	N.~V. Chawla, K.~W. Bowyer, L.~O. Hall, and W.~P. Kegelmeyer, ``{SMOTE:
		Synthetic Minority Over-Sampling Technique},'' \emph{Journal of Artificial
		Intelligence Research}, vol.~16, pp. 321--357, 2002.
	
	\bibitem{scikit-learn}
	F.~Pedregosa, G.~Varoquaux, A.~Gramfort, V.~Michel, B.~Thirion, O.~Grisel,
	M.~Blondel, P.~Prettenhofer, R.~Weiss, V.~Dubourg, J.~Vanderplas, A.~Passos,
	D.~Cournapeau, M.~Brucher, M.~Perrot, and E.~Duchesnay, ``Scikit-learn:
	Machine learning in {P}ython,'' \emph{Journal of Machine Learning Research},
	vol.~12, pp. 2825--2830, 2011.
	
	\bibitem{accuracy_paradox}
	F.~J. Valverde-Albacete and C.~Pel{\'{a}}ez-Moreno, ``{100{\%} classification
		accuracy considered harmful: The normalized information transfer factor
		explains the accuracy paradox},'' \emph{PLoS ONE}, vol.~9, no.~1, 2014.
	
	\bibitem{Tsai_intrusion_detection_machine_learning_review}
	C.-F. Tsai, Y.-F. Hsu, C.-Y. Lin, and W.-Y. Lin, ``{Intrusion detection by
		machine learning: A review},'' pp. 11\,994--12\,000, 2009.
	
	\bibitem{Saez_SMOTE_noisy_borderline}
	J.~A. Sez, J.~Luengo, J.~Stefanowski, and F.~Herrera, ``{SMOTE-IPF: Addressing
		the noisy and borderline examples problem in imbalanced classification by a
		re-sampling method with filtering},'' \emph{Information Sciences}, vol. 291,
	no.~C, pp. 184--203, 2015.
	
	\bibitem{class_imbalance_problem}
	X.~Guo, Y.~Yin, C.~Dong, G.~Yang, and G.~Zhou, ``{On the class imbalance
		problem},'' in \emph{Proceedings - 4th International Conference on Natural
		Computation, ICNC 2008}, vol.~4, 2008, pp. 192--201.
	
	\bibitem{RandomForest_Breiman}
	L.~Breiman, ``{Random Forests},'' \emph{Machine Learning}, vol.~45, no.~1, pp.
	5--32, 2001.
	
	\bibitem{SMOTEBoost}
	N.~V. Chawla, A.~Lazarevic, L.~O. Hall, and K.~W. Bowyer, ``{SMOTEBoost:
		Improving Prediction of the Minority Class in Boosting},'' 2003, pp.
	107--119.
	
	\bibitem{Rish_naive_bayes_properties}
	I.~Rish, ``{An empirical study of the naive Bayes classifier},'' \emph{IJCAI
		2001 Workshop on Empirical Methods in Artificial Intelligence}, vol.~3,
	no.~22, pp. 41--46, 2001.
	
	\bibitem{Han_Borderline_SMOTE}
	H.~Han, W.-Y. Wang, and B.-H. Mao, ``{Borderline-SMOTE: a new oversampling
		method in imbalanced data sets learning},'' \emph{LNCS}, vol. 3644, pp.
	878--887, 2005.
	
	\bibitem{Liu_undersampling}
	X.-Y. Liu, J.~Wu, and Z.-H. Zhou, ``{Exploratory undersampling for
		class-imbalance learning},'' \emph{IEEE Transactions on Systems, Man and
		Cybernetics}, vol.~39, no.~2, pp. 539--550, 2009.
	
	\bibitem{galar_ensembles_for_imbalanced_classes}
	M.~Galar, A.~Fernandez, E.~Barrenechea, H.~Bustince, and F.~Herrera, ``{A
		review on ensembles for the class imbalance problem: bagging-, boosting-, and
		hybrid-based approaches},'' pp. 463--484, 2012.
	
	\bibitem{Wang_et_al}
	G.~Wang, J.~Hao, J.~Ma, and L.~Huang, ``{A new approach to intrusion detection
		using Artificial Neural Networks and fuzzy clustering},'' \emph{Expert
		Systems with Applications}, vol.~37, no.~9, pp. 6225--6232, 2010.
	
	\bibitem{Kayacik_SOM}
	H.~G. Kayacik, A.~N. {Zincir-Heywood}, and M.~I. Heywood, ``{A hierarchical
		SOM-based intrusion detection system},'' \emph{Engineering Applications of
		Artificial Intelligence}, vol.~20, no.~4, pp. 439--451, 2007.
	
	\bibitem{Abadeh_Fuzzy_GA}
	M.~S. Abadeh, J.~Habibi, Z.~Barzegar, and M.~Sergi, ``{A parallel genetic local
		search algorithm for intrusion detection in computer networks},''
	\emph{Engineering Applications of Artificial Intelligence}, vol.~20, no.~8,
	pp. 1058--1069, 2007.
	
\end{thebibliography}
%


\end{document}